\documentclass[10pt, conference, compsocconf]{IEEEtran}
\ifCLASSINFOpdf
  \usepackage[pdftex]{graphicx}
\else
\fi
\usepackage[cmex10]{amsmath}
\usepackage{subcaption}
\usepackage[hyperfootnotes=false]{hyperref}

\begin{document}
%
\title{FaSTExt: Fast and Small Text Extractor}




%
\author{\IEEEauthorblockN{Alexander Filonenko\IEEEauthorrefmark{1},
Konstantin Gudkov\IEEEauthorrefmark{1},
Aleksei Lebedev\IEEEauthorrefmark{1}, 
Ivan Zagaynov\IEEEauthorrefmark{1}, and
Nikita Orlov\IEEEauthorrefmark{2}}
\IEEEauthorblockA{\IEEEauthorrefmark{1}ABBYY\\
Moscow, Russia\\ Email: \{alexander.filonenko, konstantin\_g, aleksey.lebedev, ivan.zagaynov\}@abbyy.com}
\IEEEauthorblockA{\IEEEauthorrefmark{2}Moscow Institute of Physics and Technology\\
Moscow, Russia\\
Email: nikita.orlov@phystech.edu}
}


\maketitle

\begin{abstract}
Text detection in natural images is a challenging but necessary task for many applications. Existing approaches utilize large deep convolutional neural networks making it difficult to use them in real-world tasks. We propose a small yet relatively precise text extraction method. The basic component of it is a convolutional neural network which works in a fully-convolutional manner and produces results at multiple scales. Each scale output predicts whether a pixel is a part of some word, its geometry, and its relation to neighbors at the same scale and between scales. The key factor of reducing the complexity of the model was the utilization of depthwise separable convolution, linear bottlenecks, and inverted residuals. Experiments on public datasets show that the proposed network can effectively detect text while keeping the number of parameters in the range of 1.58 to 10.59 million in different configurations.

\end{abstract}


%
\IEEEpeerreviewmaketitle

\section{Introduction}
Extraction of text in natural scenes is one of the hot issues nowadays. The main limitation of utilizing existing approaches on real devices is their complexity which leads to slow processing.

We propose a fast and small network FaSTExt for extracting text from images. We focus on the scenarios where the text was acquired from a natural scene by a camera. In such a case, the text will appear close to the center of the image making it easier to create a compact text extraction system.

The main features of the proposed method are:
\begin{itemize}
    \item The network detects local parts of the words and connections between them to combine parts of the long words
    \item Prediction is done in 5 scales to prevent losing small and large symbols
    \item Inverted residual blocks are the key element in decreasing the complexity of the network
     \item A "thin" version of the network with fewer channels in each layer can still achieve appropriate performance while working fast.
\end{itemize}


\section{Related Work}
Since the implementation of deep learning became practical, text detection techniques are based on neural networks. A deep learning based method \cite{Zhang16} uses fully convolutional network (FCN) to find a probability that pixels belong to a text area. After applying maximally stable extremal regions (MSER), a shortened FCN was utilized to acquire the character centroids and with the help of intensity and geometric criteria remove false candidates.

Shi \emph{et al.} in \cite{SegLink} proposed to find segments of words and connections between them. The whole detection process of segments and links was done in a single pass of a CNN named SegLink in a fully-convolutional manner with depth-first search (DFS) and bounding box creation postprocessing. 

Zhou \emph{et al.} proposed a similar strategy in \cite{East} where a variety of postprocessing steps were eliminated by performing most of the calculations in a single U-Net-like \cite{unet} FCN named EAST which outputs word box parameters by itself. Results of computations are filtered by non-maximum suppression (NMS) and thresholding. The length of the word to be detected is limited by a receptive field of output pixels.

An ArbiText network \cite{ArbiText} based on the Single Shot Detector (SSD) applies the circle anchors to replace bounding boxes which should be more robust to orientation variations. Authors also applied pyramid pooling to preserve low-level features in deeper layers.

Liao \emph{et al.} in \cite{Liao18} applied FCN similar to SSD with VGG backbone. A key feature of their architecture was adding a dense prediction part which contains a feature map for classification insensitive to text orientation as well as a feature map for regression with rotation-sensitive features. 

An instance transformation network was proposed in \cite{Wang18} based on VGG with three outputs for each pixel location: the probability of text, parameters of the affine transformation matrix, and coordinate offsets. The network is able to detect text in multiple scales and orientations.

Liu \emph{et al.} in \cite{FOTS} combined text detection and recognition parts in one end-to-end CNN. The backbone of the network is the Feature Pyramid Network which incorporates residual operations from ResNet-50 \cite{Resnet}. The network, in text detection part, outputs text probability, bounding box distances in four directions, and a rotation angle of the bounding box. The smallest real-time version contains 29 million parameters.

\section{Methodology}
The main goal of creating a new method for text extraction in natural scenes was the lack of small but precise and fast deep neural networks which can work in the hardware constrained environment.  

Detecting a whole word by applying a set of predefined bounding boxes is not the most efficient approach since width to height ratio of the words can vary over a wide range.

\begin{figure}[!t]
\centering
\includegraphics[width=3.3in]{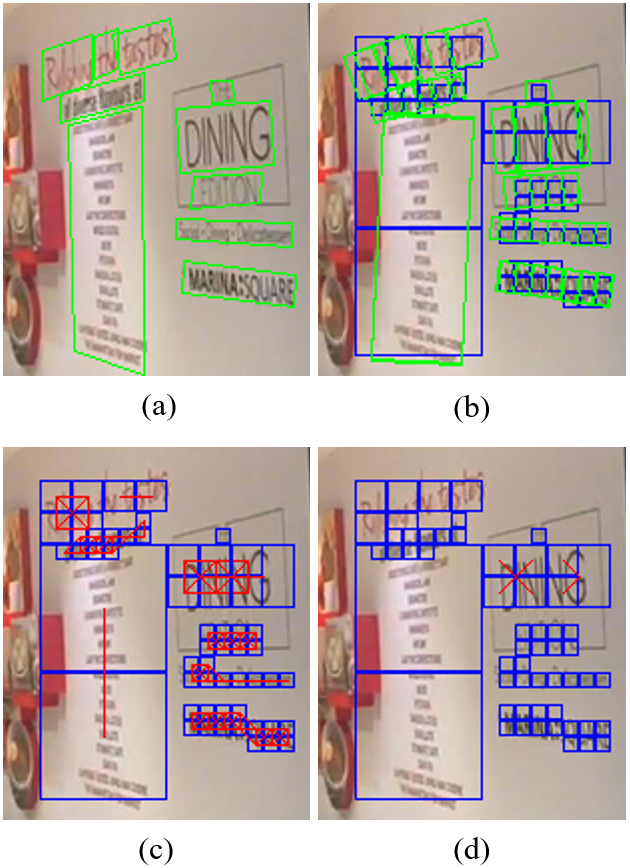}
\caption{Example of segments and links. Figure (a) illustrates the original markup as green boxes around each word. Blue squares in the figure (b) are the receptive fields of output pixels at the different scales while green boxes are segments which are shifted, scaled, and oriented boxes each related to their output pixel. Red lines in figure (c) are the links between segments belonging to the same word. Red lines in figure (d) show cross-layer links of between different scales which occupy the same part of the word. }
\label{fig_seg_links}
\end{figure}

In our approach, we follow the idea proposed in \cite{SegLink} by detecting a local part of the word named segment which is essentially an oriented rectangle which follows the direction of the word. Segments belonging to the same word are connected via links to their neighbors in eight directions in the same scale and to the segments in the smaller scale (cross-layer links). Segments and links are computed by a single CNN in a fully-convolutional manner. The general idea of segments and links is illustrated in Fig.~\ref{fig_seg_links}. Separate words are outlined by quadrangles (green boxes in  Fig.~\ref{fig_seg_links}a). Each pixel of the network has a rectangular receptive field (blue squares in Fig.~\ref{fig_seg_links}b) with the size determined by the number of strides between the input of the network and the current output. Green quadrangles in Fig.~\ref{fig_seg_links}b are the segments. Each segment is shifted from the center of the receptive field of the respected pixel, scaled, and rotated to follow the direction of the whole word. Links and cross-layer links are shown as red lines in Fig.~\ref{fig_seg_links}c and Fig.~\ref{fig_seg_links}d respectively.

It is expected for the proposed network to work within a limited range of resolutions to guarantee real-time performance. SegLink used six scales in which text was detected while our experiments have shown that extracting text in five scales at approximately 0.3 megapixels was a rational tradeoff between accuracy and processing time. The range of scales is [$8\times8$, $16\times16$, $32\times32$, $64\times64$, $128\times128$].

The original SegLink and most of the text detection networks utilize VGG as a backbone which in our truncated implementation contains 23 million parameters. A large number of parameters leads to an increased quantity of mathematical operations required in inference.

Sandler \emph{et al.} introduced their MobileNetV2 \cite{Mobilenetv2} specifically designed to operate in hardware constrained conditions while keeping detection performance similar to VGG. The basic parts of the approach are:
\begin{itemize}
    \item Replace full convolutional operator by a depthwise separable convolution effectively reducing computations;
    \item Use linear bottlenecks (without activations) to prevent losing too much information;
    \item Use shortcuts between bottlenecks which are small but contain all the information in a compressed state.
\end{itemize}

FaSTExt utilizes bottleneck blocks of MobileNetV2 as the basic building structure. The full structure of FaSTExt is shown in Table~\ref{table_model}. Note that the actual number of channels for each layer depends on multiplier $\alpha$.

Sandler \emph{et al.} used dropout and BatchNorm during training. FaSTExt contains only Batchnorms as dropout may have a negative impact on convergence as discussed in \cite{res}.

The input of the network is expected to be a three-channel image of arbitrary size. All the convolutions in FaSTExt are performed with a $3\times3$ kernel to preserve fast processing speed. Unlike MobileNetV2, FaSTExt refines initial features by a depthwise convolution operation with fewer channels before the first bottleneck to compress the data.

The very first output of the network contains two extra bottleneck blocks placed before it which detached from the main network graph and thus do not share weights with other scales. Without such extra block, the number of layers leading to the first output was not sufficient, and segments with $8\times8$ receptive field resulted in lower prediction accuracy than the deeper ones.

Each output of the network (except the first one) contains 31 channels as shown in Table~\ref{table_outputs}. The first two channels are text/non-text class predictions of the pixel. Next five channels contain information to restore segment location: shifts in vertical and horizontal directions, change in width and height, and rotation. The following sixteen channels describe the connection to eight neighboring segments. The last eight channels describe cross-layer links to four segments in the previous scale.  Every two channels of within and cross-layer links are filtered by softmax to get the score of each link. The output for the very first scale does not contain channels for cross-layer links. 

The network produces segments and links, not the bounding box of the word. To build the latter one a DFS is used to find connected components where segments and links are nodes and edges of the graph respectively. Segments connected by links are used to build the final bounding box via computing their average angle and fitting a line to segments centers rotated by this angle. The farthest projected centers to the line will be the boundaries of the word. The original bounding box reconstruction algorithm is shown as Algorithm 1 in \cite{SegLink}.

\begin{table}[!t]
\renewcommand{\arraystretch}{1.3}
\caption{FaSTExt. Each line describes a layer with \emph{c} output channels and \emph{s} strides. All spatial convolutions use $3\times3$ kernels.  Some layers have extra bottleneck blocks which lead to a specific output. \emph{Scale} shows that a layer leads to output with \emph{Scale} receptive field for each pixel.  Note that the actual number of output channels in bottlenecks will be adjusted in experiments by $\alpha$ leading to \emph{c}$\times\alpha$ channels.}
\label{table_model}
\centering
\begin{tabular}{|l||c||c||c||c||c|}
\hline 
\textbf{Operator} & \textbf{c} & \textbf{s} & \textbf{Extra blocks} &  \textbf{Scale} \\ 
\hline 
conv2d & 32 & 2 & - & - \\ 
\hline 
depthwise & 16 & 1 & - & - \\ 
\hline 
bottleneck1 & 24 & 2 & - & - \\ 
\hline 
bottleneck2 & 24 & 1 & - & - \\ 
\hline 
bottleneck3 & 32 & 2 & - & - \\ 
\hline 
bottleneck4 & 32 & 1 & - & - \\ 
\hline 
bottleneck5 & 32 & 1 & 2 & 8 \\ 
\hline 
bottleneck6 & 64 & 2 & - & - \\ 
\hline 
bottleneck7 & 64 & 1 & - & - \\ 
\hline 
bottleneck8 & 64 & 1 & - & - \\ 
\hline 
bottleneck9 & 64 & 1 & - & - \\ 
\hline 
bottleneck10 & 96 & 1 & - & - \\ 
\hline 
bottleneck11 & 96 & 1 & - & - \\ 
\hline 
bottleneck12 & 96 & 1 & - & 16 \\ 
\hline 
bottleneck13 & 140 & 2 & - & - \\ 
\hline 
bottleneck14 & 140 & 1 & - & - \\ 
\hline 
bottleneck15 & 140 & 1 & - & - \\ 
\hline 
bottleneck16 & 140 & 1 & - & 32 \\ 
\hline 
bottleneck17 & 140 & 2 & - & - \\ 
\hline 
bottleneck18 & 140 & 1 & - & - \\ 
\hline 
bottleneck19 & 140 & 1 & - & 64 \\ 
\hline 
bottleneck20 & 140 & 2 & - & - \\ 
\hline 
bottleneck21 & 140 & 1 & - & 128 \\ 
\hline 

\end{tabular}
\end{table}

\begin{table}[!t]
\renewcommand{\arraystretch}{1.3}
\caption{The structure of the output layers. Each layer consists of 31 channels except for the very first scale which has no cross-layer links.}
\label{table_outputs}
\centering
\begin{tabular}{|c||c||c|}
\hline 
\textbf{Role} & \textbf{Number of channels}\\ 
\hline 
Class labels & 2\\ 
\hline 
Cell geometry & 5 \\ 
\hline 
Links & 16 \\ 
\hline 
Cross-layer links & 8 \\ 
\hline 
\end{tabular}
\end{table}

\section{Experiments}

\subsection{Benchmark Datasets}
All the benchmark datasets were chosen to reflect the camera-based scenario of text detection.
\subsubsection{Pretraining}
The network was pretrained on a large dataset SynthText in the Wild \cite{synthtext} with more than 800,000 images. The pretraininig process was performed for 20 epochs without augmentation and took 20 days for each $\alpha$. 

\subsubsection{ICDAR 2013}
ICDAR2013 (IC13) \cite{ic13} is a small dataset with 229 training and 233 testing images. The text is centered in many images which reflects well the scenario of camera-based text detection.

\subsubsection{ICDAR 2015}
ICDAR 2015 (IC15) \cite{ic15} is a more challenging dataset than IC13. In many cases, a text is situated at the edge of the image and resides in shadows. IC15 consists of 1000 and 500 training and testing images respectively.

\subsubsection{MSRA-TD500}
MSRA-TD500 (TD500) \cite{td500} includes pictures of natural scenes with rotated text and multiple languages with 300 training and 200 testing images.

\subsubsection{Special cases}
IC15 and MSRA-TD500 contain bounding boxes marked as "do not care" and "difficult". We used these boxes during training to help the model achieve better generalization. However, in the evaluation, these special entities were not penalizing the accuracy.

\subsection{Implementation Details}
Experiments were conducted on machines with Nvidia GTX 1080Ti designated for each training routine. Implementation is done using Keras framework and Python 3.6.

Training on each dataset took a various number of epochs, and we stopped training when loss reached a plateau.

The thresholds for segments and links were set to 0.5 for all tests. It was reported in \cite{SegLink} that changing these thresholds does not affect the result drastically. The evaluation is done per ground truth quadrangle and detected one, not per image.

Cross-entropy has been used as the loss for classification of segments and links. Huber loss has been utilized for computing the geometry of cells.
As the optimizer, we used the AMSGrad version of Adam proposed in \cite{adam}. ReLU6 was used as an activation function.

We follow the evaluation protocol of the Robust Reading Competition\footnote{\url{http://rrc.cvc.uab.es/?ch=2\&com=tasks}} which considers a detection as a match if it overlaps a ground truth bounding box by more than 50\%. Some datasets mark every word separately, others may include several words in the ground truth box. To reflect possible variations, we combine three types of detections:
\begin{itemize}
    \item one-to-one where one ground truth box matches a single detection box;
    \item one-to-many where multiple detection boxes together match a single ground truth box;
    \item many-to-one where a single detection box matches several ground-truth boxes.
\end{itemize}

In all tests, the images were resized to have 512 pixels at the smallest size while keeping the original aspect ratio in order to avoid distortion of the text. This relatively small resolution leads to lower detection performance on datasets with small texts; however, in the real use scenario, it is expected that a camera will be pointed to a text area thus keeping words close to the center of the image with appropriate scale.

\subsubsection{Hard Negative Mining}
Usually, the text covers a small portion of the image leading to a highly unbalanced training set. As in \cite{SegLink} we select the hard negative samples; however, the loss $L$ for non-hard samples is not set to 0. A combined loss for class labels, links, and cross-layer links is computed by:

\begin{equation}\label{eq:loss}
L = \dfrac{1}{P_{t}}L_{p} + \dfrac{1}{N_{t}}L_{n} + \dfrac{2}{3N_{h}}L_{h},
\end{equation}
where $N_h$, $N_t$, and  $P_t$ is the number of hard negative, negative, and positive samples respectively. 

Hard examples are chosen by sorting the output pixel values by loss and taking the $N_h$ of them for which the network works worst. The number of hard negatives $N_h$ is computed by:

\begin{equation}\label{eq:ohem}
N_{h} = min\left( N_{t}, max \left( 10, 2P_{t} \right)   \right).
\end{equation}

\subsubsection{Augmentation}
Since public datasets for text detection are small for proper training of deep neural networks, we randomly apply the following extensive augmentation to the train sets: Gaussian blur, average blur, median blur, sharpen filter, emboss, simplex noise, additive Gaussian noise, dropout, invert colors, shift hue, saturation, and value, change contrast, convert to grayscale, elastic transformation, rotate, randomly crop, flip, perspective distortion.

\subsection{Experimental Results}
To choose the appropriate model suitable for our scenario of extracting text from images taken by cameras we tested three variations of the FaSTExt by applying three different "thickness multiplier" $\alpha$ values: 0.75, 1, and 2. 

Table~\ref{table_num_parameters} shows the number of parameters FaSTExt has with different $\alpha$. The smallest version of FaSTExt achieved 37.5 frames per second (FPS) performance on average as shown in Table~\ref{table_time} when used Nvidia GTX 1080Ti. A similar network with segments and links \cite{SegLink} maxed out at 16.5 FPS in our implementation. EAST with Resnet-50 backbone shows 15.6 FPS performance and is slightly slower than the "thickest" FaSTExt. There is a very small difference in processing time between $\alpha$=0.75 and $\alpha$=1. The largest FaSTExt is still fast with 18.2 FPS but it should be chosen only for scenarios where it performs considerably better than lighter versions. 
\begin{table}[!t]
\renewcommand{\arraystretch}{1.3}
\caption{Number of parameters with different values of $\alpha$ multiplier.}
\label{table_num_parameters}
\centering
\begin{tabular}{|c||c|}
\hline 
\textbf{$\alpha$} & \textbf{Million parameters}\\ 
\hline 
0.75 & 1.58 \\ 
\hline 
1.00 & 2.87 \\ 
\hline 
2.00 & 10.59 \\ 
\hline 
\end{tabular}
\end{table}

\begin{table}[!t]
\renewcommand{\arraystretch}{1.3}
\caption{Processing time on Nvidia GTX1080 Ti.}
\label{table_time}
\centering
\begin{tabular}{|c||c||c|}
\hline 
\textbf{Model} & \textbf{Time, ms} & \textbf{Frames per second}\\ 
\hline 
\textbf{FaSTExt} ($\alpha$=0.75) & 26.7 & 37.5 \\ 
\hline 
\textbf{FaSTExt} ($\alpha$=1.00) & 28.4 & 35.2 \\ 
\hline 
\textbf{FaSTExt} ($\alpha$=2.00) & 55.0 & 18.2 \\ 
\hline 
EAST+ResNet-50 & 64.0 & 15.6 \\ 
\hline 
SegLink (VGG) & 60.4 & 16.5 \\ 
\hline 
\end{tabular}
\end{table}

\begin{table}[!t]
\renewcommand{\arraystretch}{1.3}
\caption{Comparison of detection results on ICDAR 2013 dataset. MS means multi-scale detection. $\bullet$ means "no data". OI means "our implementation".}
\label{table_ic13}
\centering
\begin{tabular}{|c||c||c||c|}
\hline 
\textbf{Method} & \textbf{Precision} & \textbf{Recall} & \textbf{F-measure} \\ 
\hline 
Zhang \emph{et al.} \cite{Zhang16} & 0.88 & 0.78 & 0.83 \\ 
\hline 
Shi \emph{et al.} \cite{SegLink} & 0.877 & 0.83 & 0.853 \\ 
\hline 
Liao \emph{et al.} \cite{Liao18} & 0.88 & 0.75 & 0.81 \\ 
\hline 
Liao \emph{et al.} \cite{Liao18} MS & 0.92 & 0.86 & 0.89 \\ 
\hline 
Liu \emph{et al.} \cite{FOTS} & $\bullet$ & $\bullet$ & 0.8823 \\ 
\hline 
Liu \emph{et al.} \cite{FOTS} MS & $\bullet$ & $\bullet$ & 0.925 \\ 
\hline 
EAST+ResNet-50 OI & 0.8335 & 0.7452 & 0.7869 \\ 
\hline 
\textbf{FaSTExt} ($\alpha$=0.75) & 0.8774 & 0.8659 & 0.8716 \\ 
\hline
\textbf{FaSTExt} ($\alpha$=1.00) & 0.8693 & 0.8838 & 0.8765 \\ 
\hline 
\textbf{FaSTExt} ($\alpha$=2.00) & 0.8774 & 0.9174 & 0.8970 \\ 
\hline 
\end{tabular}
\end{table}

Detection performance of FaSTExt on IC13 is shown in Table~\ref{table_ic13}. Our network with $\alpha$=2 outperforms the single scale versions of state-of-the-art networks. Smaller versions of FaSTExt show the detection performance on par with the state of the art. Multiscale implementations of existing networks extract text with higher accuracy but require much more computations. Zhou \emph{et al.} \cite{East} did not report detection performance of their approach on IC13. Our implementation of their approach does not outperform FaSTExt.

Detection performance of FaSTExt on IC15 is shown in Table~\ref{table_ic15}. The placement and appearance of the text are more difficult comparing to IC13. FaSTExt does not achieve the accuracy of its much heavier counterparts. Through this part of the experiment the limitations of our proposed network can be found in comparing IC15 to IC13: FaSTExt requires the text area to be located closer to the center of the image, the size of the text should not be very small, and it should not be rotated much.

TD500 was created to be more challenging than IC13, and it can be seen in Table~\ref{table_td500} that although FaSTExt shows 14\% higher F-measure here, it still lags behind the state of the art in terms of detection quality.

\begin{figure*}
    \centering
    \begin{subfigure}[b]{0.25\linewidth}%
        \includegraphics[width=0.98\textwidth]{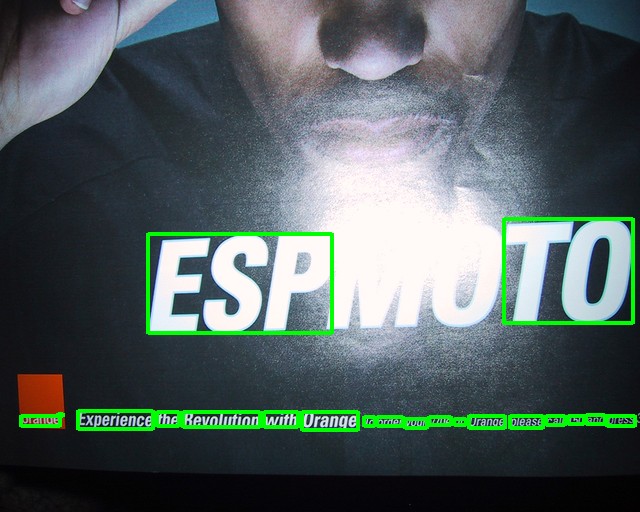}
        \label{fig:ex_gt_1}
    \end{subfigure}%
    \begin{subfigure}[b]{0.25\linewidth}%
        \includegraphics[width=0.98\textwidth]{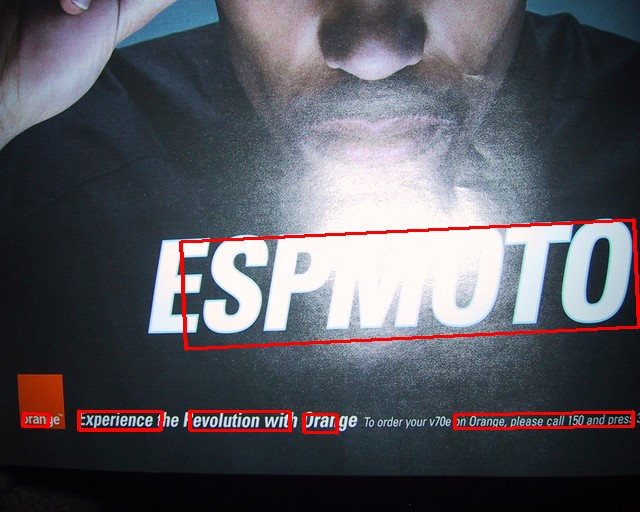}
        \label{fig:ex_r_075}
    \end{subfigure}%
    \begin{subfigure}[b]{0.25\linewidth}%
        \includegraphics[width=0.98\textwidth]{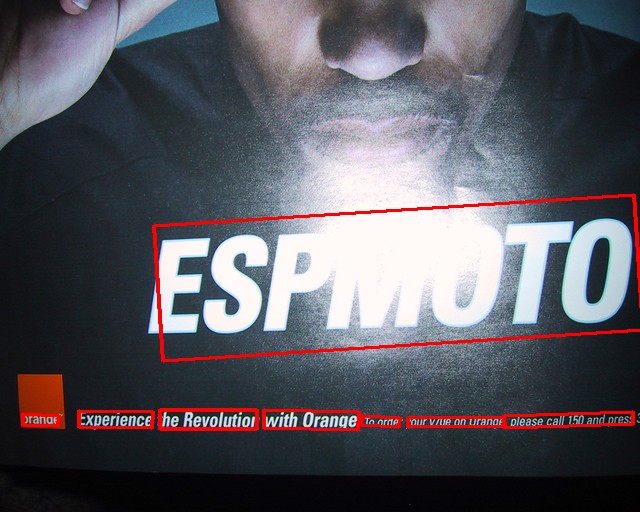}
        \label{fig:ex_r_100}
    \end{subfigure}%
    \begin{subfigure}[b]{0.25\linewidth}%
        \includegraphics[width=0.98\textwidth]{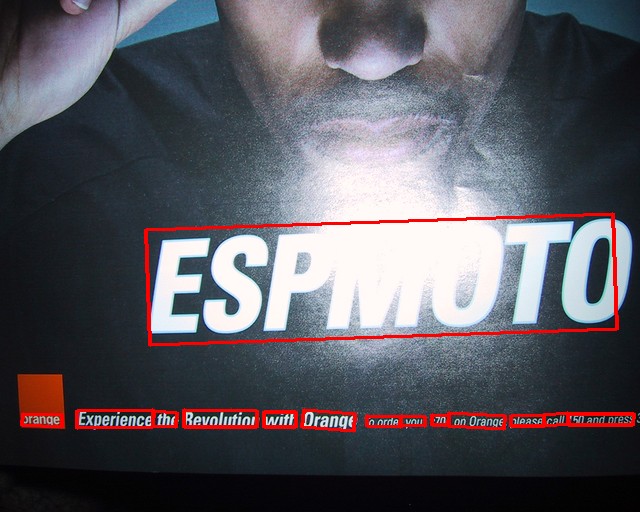}
        \label{fig:ex_r_200}
    \end{subfigure}%
  
    \begin{subfigure}[b]{0.25\linewidth}%
        \includegraphics[width=0.98\textwidth]{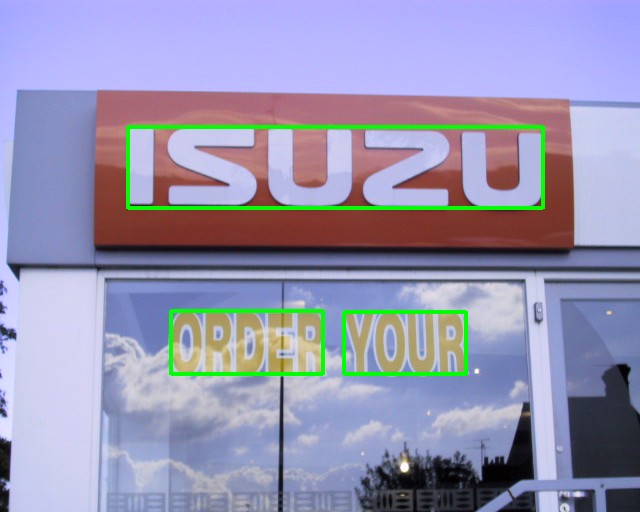}
        \label{fig:ex_gt_1}
    \end{subfigure}%
    \begin{subfigure}[b]{0.25\linewidth}%
        \includegraphics[width=0.98\textwidth]{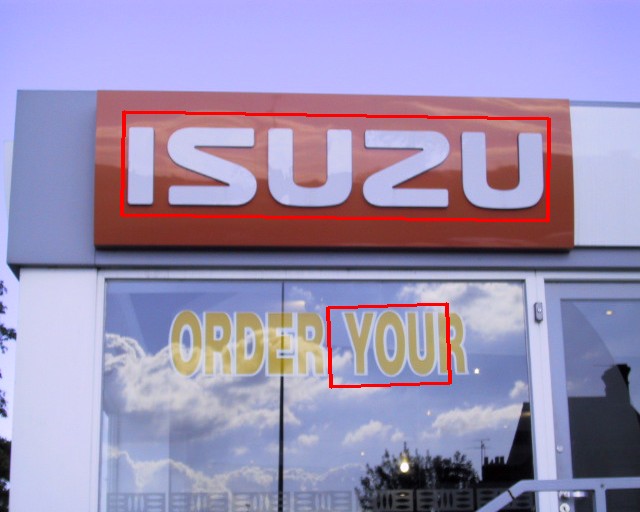}
        \label{fig:ex_r_075}
    \end{subfigure}%
    \begin{subfigure}[b]{0.25\linewidth}%
        \includegraphics[width=0.98\textwidth]{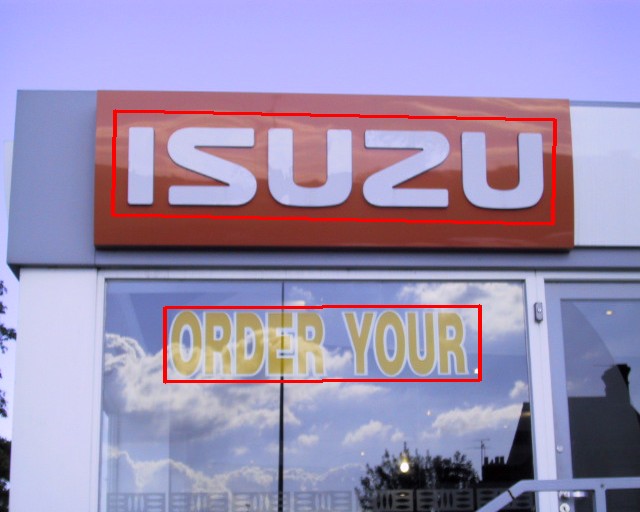}
        \label{fig:ex_r_100}
    \end{subfigure}%
    \begin{subfigure}[b]{0.25\linewidth}%
        \includegraphics[width=0.98\textwidth]{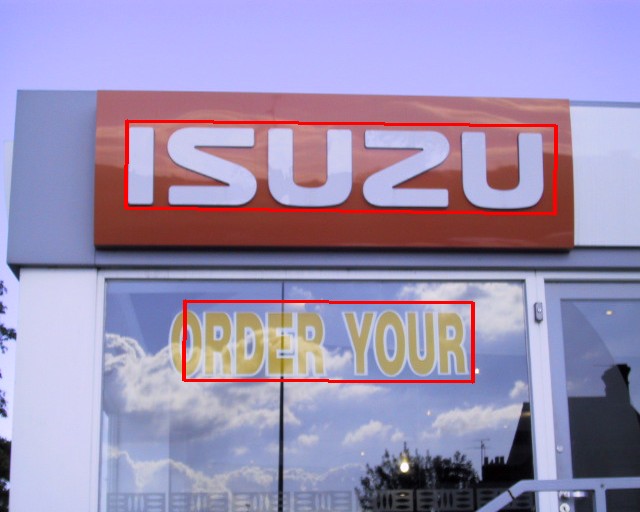}
        \label{fig:ex_r_200}
    \end{subfigure}%
    
    \begin{subfigure}[b]{0.25\linewidth}%
        \includegraphics[width=0.98\textwidth]{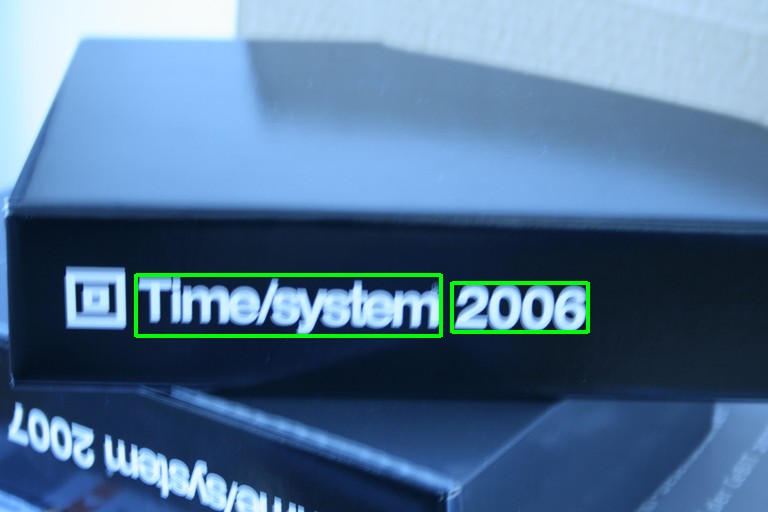}
        \caption{Ground truth}
        \label{fig:ex_gt_1}
    \end{subfigure}%
    \begin{subfigure}[b]{0.25\linewidth}%
        \includegraphics[width=0.98\textwidth]{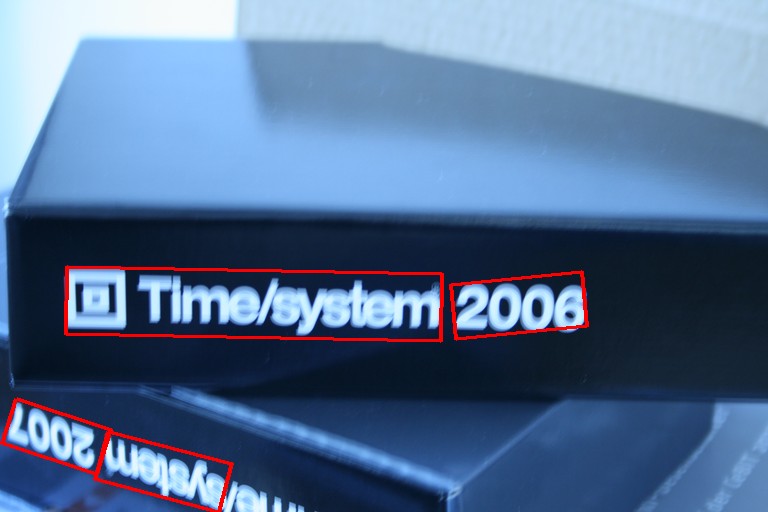}
        \caption{$\alpha$=0.75}
        \label{fig:ex_r_075}
    \end{subfigure}%
    \begin{subfigure}[b]{0.25\linewidth}%
        \includegraphics[width=0.98\textwidth]{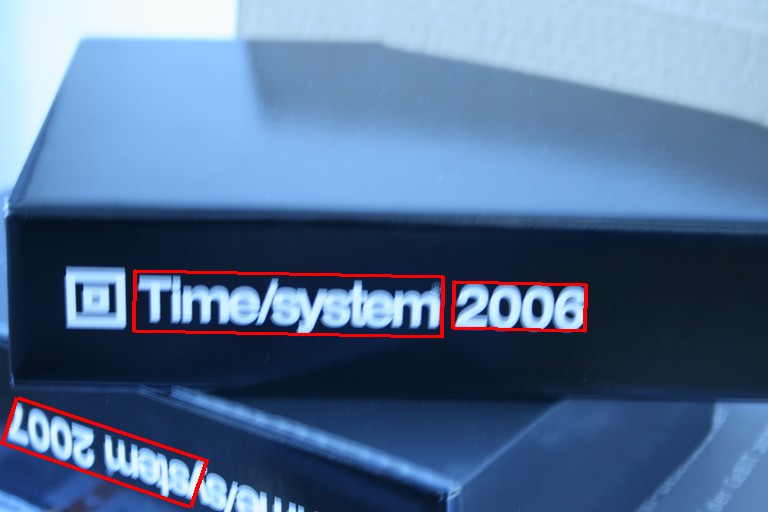}
        \caption{$\alpha$=1}
        \label{fig:ex_r_100}
    \end{subfigure}%
    \begin{subfigure}[b]{0.25\linewidth}%
        \includegraphics[width=0.98\textwidth]{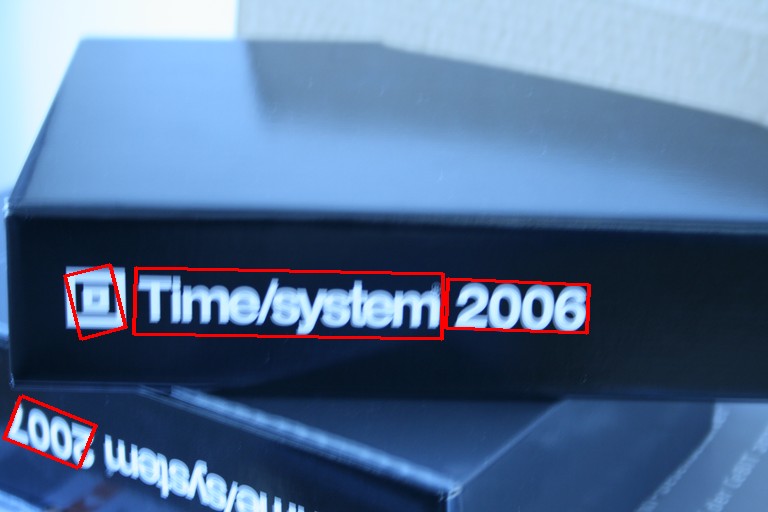}
        \caption{$\alpha$=2}
        \label{fig:ex_r_200}
    \end{subfigure}%
     \caption{Example of text extraction results. The first column contains ground truth which accompanied the dataset. Second to fourth columns show results with  $\alpha$ values 0.75, 1, and 2 respectively.}\label{fig:ex}
\end{figure*}

Example of text extraction with different $\alpha$ values is illustrated in Fig.~\ref{fig:ex}. In the first row, ground truth markup didn't include the glared area while FaSTExt could successfully detect the full word at this difficult condition. Some small words in the bottom of the image were not detected by FaSTExt with $\alpha$=0.75 when two other networks succeeded. A similar situation has happened for the image in the second row. The difference in F-measure between $\alpha$=0.75 and $\alpha$=1 for IC13 is very small but the quality of detection is evidently better in the latter case. The markup for the third image missed the reflected text in the bottom increasing confusion for the network during training. Higher $\alpha$ does not improve the detection performance on IC15 and TD500, and FaSTExt with $\alpha$=0.75 works better. The possible reason for this is that the more filters the network has the more data it is required to train the network so it can achieve the good generalization. All test datasets are rather small, but easier detection conditions of IC13 help finding correct weights for the convolutional kernels.

In general, experiments have shown that FaSTExt achieves high accuracy when working in a controlled environment where a user points one's device to the text area and correctly orients it.

\begin{table}[!t]
\renewcommand{\arraystretch}{1.3}
\caption{Comparison of detection results on ICDAR 2015 dataset. MS means multi-scale detection.}
\label{table_ic15}
\centering
\begin{tabular}{|c||c||c||c|}
\hline 
\textbf{Method} & \textbf{Precision} & \textbf{Recall} & \textbf{F-measure} \\ 
\hline 
Zhang \emph{et al.} \cite{Zhang16} & 0.71 & 0.43 & 0.54 \\ 
\hline 
Shi \emph{et al.} \cite{SegLink} & 0.731 & 0.768 & 0.75 \\ 
\hline 
Zhou \emph{et al.} \cite{East} & 0.8357 & 0.7347 & 0.782 \\ 
\hline 
Xing \emph{et al.} \cite{ArbiText} & 0.792 & 0.735 & 0.759 \\ 
\hline 
Liao \emph{et al.} \cite{Liao18} & 0.856 & 0.79 & 0.822 \\ 
\hline 
Liao \emph{et al.} \cite{Liao18} MS & 0.88 & 0.8 & 0.838 \\ 
\hline 
Liu \emph{et al.} \cite{FOTS} & 0.91 & 0.8517 & 0.8799 \\ 
\hline 
Liu \emph{et al.} \cite{FOTS} MS & 0.9185 & 0.8792 & 0.8984 \\ 
\hline 
Liu \emph{et al.} \cite{FOTS} RT & 0.8595 & 0.7983 & 0.8278 \\ 
\hline 
Wang \emph{et al.} \cite{Wang18} & 0.857 & 0.741 & 0.795 \\ 
\hline 
\textbf{FaSTExt} ($\alpha$=0.75) & 0.6501 & 0.4154 & 0.5069 \\ 
\hline 
\textbf{FaSTExt} ($\alpha$=1.00) & 0.5983 & 0.3966 & 0.4770 \\ 
\hline 
\textbf{FaSTExt} ($\alpha$=2.00) &  0.5975 & 0.4326 & 0.5019 \\ 
\hline 
\end{tabular}
\end{table}

\begin{table}[!t]
\renewcommand{\arraystretch}{1.3}
\caption{Comparison of detection results on MSRA-TD500 dataset.}
\label{table_td500}
\centering
\begin{tabular}{|c||c||c||c|}
\hline 
\textbf{Method} & \textbf{Precision} & \textbf{Recall} & \textbf{F-measure} \\ 
\hline 
Zhang \emph{et al.} \cite{Zhang16} & 0.83 & 0.67 & 0.74 \\ 
\hline 
Shi \emph{et al.} \cite{SegLink} & 0.86 & 0.7 & 0.77 \\ 
\hline 
Zhou \emph{et al.} \cite{East} & 0.8728 & 0.6743 & 0.7608 \\ 
\hline 
Xing \emph{et al.} \cite{ArbiText} & 0.78 & 0.72 & 0.75 \\ 
\hline 
Liao \emph{et al.} \cite{Liao18} & 0.87 & 0.73 & 0.79 \\ 
\hline 
Wang \emph{et al.} \cite{Wang18} & 0.903 & 0.723 & 0.803 \\ 
\hline 
\textbf{FaSTExt} ($\alpha$=0.75) & 0.7362 &  0.5754 & 0.6459 \\ 
\hline 
\textbf{FaSTExt} ($\alpha$=1.00) & 0.6179 & 0.6207 & 0.6193 \\ 
\hline 
\textbf{FaSTExt} ($\alpha$=2.00) & 0.6513 & 0.6292 & 0.6401 \\ 
\hline 
\end{tabular}
\end{table}

\section{Conclusion}
We proposed a fast and small network FaSTExt for extracting text from images. The network is application specific, and it performs fast and accurate in a scenario where text is aligned to the center of the image and not hidden by strong shadows.

In the future works, different lightweight types of convolutions will be investigated to further decrease the size of FaSTExt. The structure of the network itself will be modified to give better contexts to the smaller resolutions in order to deal with the small texts introduced by IC15 and TD500 datasets.

\bibliographystyle{IEEEtran}
\bibliography{IEEEabrv,references}
%

\end{document}